\documentclass[runningheads]{llncs}

\usepackage{eccv}

\usepackage{eccvabbrv}

\usepackage{graphicx}
\usepackage{booktabs}

\usepackage[accsupp]{axessibility}  

\usepackage{multirow}
\usepackage{booktabs}
\usepackage{makecell}
\usepackage[table]{xcolor}
\usepackage{titletoc}

\usepackage{hyperref}

\usepackage{orcidlink}

\begin{document}

\title{SignAgent: Agentic LLMs for Linguistically-Grounded Sign Language Annotation and Dataset Curation}

\titlerunning{SignAgent}

\author{Oliver Cory\orcidlink{0009-0002-5383-7202} \and
Ozge Mercanoglu Sincan\orcidlink{0000-0001-9131-0634} \and
Richard Bowden\orcidlink{0000-0003-3285-8020}}

\authorrunning{O.~Cory et al.}

\institute{Centre for Vision, Speech and Signal Processing, University of Surrey, Guildford, UK}

\maketitle
\begin{abstract}
This paper introduces \emph{SignAgent}, a novel agentic framework that utilises Large Language Models (LLMs) for scalable, linguistically-grounded Sign Language (SL) annotation and dataset curation. Traditional computational methods for SLs often operate at the gloss level, overlooking crucial linguistic nuances, while manual linguistic annotation remains a significant bottleneck, proving too slow and expensive for the creation of large-scale, phonologically-aware datasets. \emph{SignAgent} addresses these challenges through \emph{SignAgent Orchestrator}, a reasoning LLM that coordinates a suite of linguistic tools, and \emph{SignGraph}, a knowledge-grounded LLM that provides lexical and linguistic grounding. 
We evaluate our framework on two downstream annotation tasks. First, on \emph{Pseudo-gloss Annotation}, where the agent performs constrained assignment, using multi-modal evidence to extract and order suitable gloss labels for signed sequences. Second, on \emph{ID Glossing}, where the agent detects and refines visual clusters by reasoning over both visual similarity and phonological overlap to correctly identify and group lexical sign variants. Our results demonstrate that our agentic approach achieves strong performance for large-scale, linguistically-aware data annotation and curation.

\end{abstract}
    
\section{Introduction}
\label{sec:intro}

Sign Languages (SLs) are rich visual–gestural languages whose structure is expressed through coordinated manual and non-manual phonological components, including handshape, movement, location, orientation, and facial cues. Developing large-scale SL resources requires an understanding of how these components combine into meaningful lexical units. Yet current computational SL research remains bottlenecked by the absence of systems capable of performing \textit{linguistic reasoning} over these multimodal signals. 
The manual creation of detailed linguistic annotation is prohibitively expensive and time-consuming, with one minute of SL data requiring over an hour to annotate, even at the most superficial level~\cite{battisti2024automatic, woll2022segmentation}. This presents a significant barrier to scaling SL technologies that rely on phonologically labelled data.

In this work, we present \emph{SignAgent}, an agentic framework for SL annotation and dataset curation that leverages linguistic knowledge for grounded reasoning. At its core lies the \emph{SignAgent Orchestrator}, a reasoning Large Language Model (LLM) responsible for multistage decision making, tool coordination, and interaction with \emph{SignGraph}, a knowledge-grounded LLM with lexical and linguistic graphs. Our \emph{Toolset} spans phonological, syntactic, and semantic analysis of sign language video, enabling the Orchestrator to decompose complex tasks, invoke enhanced modules, and perform linguistically informed annotation through multistage reasoning.
We evaluate our framework on two key downstream annotation tasks. The first is a \emph{Pseudo-gloss Annotation task}, where the agent uses \emph{Enhanced Tools} to gather multi-modal evidence to extract and order pseudo-gloss candidates given an input sample and sentence translation. Second, we demonstrate the framework's ability to perform \textit{ID-Glossing}\footnote{A gloss is a written label representing a sign or lexical unit. ID-glossing refers to different forms of the same lexical sign, which may vary in handshape, movement, or location. For example, the sign for basketball can be produced with one hand or both hands; each production is labeled as a separate variant (e.g., basketball-1, basketball-2).}
an agentic workflow that detects and refines visual clusters by reasoning over both visual similarity and phonological overlap to correctly identify and group lexical sign variants for a given gloss.
Our results demonstrate that this agentic approach achieves strong performance in pseudo-gloss alignment, and significantly improves clustering quality for lexical variant identification, proving its effectiveness for large-scale, linguistically-aware video annotation.

In summary, our contributions are as follows: (i) We introduce the first application of agentic reasoning for SL annotation and dataset curation, combining tool-augmented multimodal evidence with knowledge-grounded retrieval. (ii) We evaluate on two complementary tasks--pseudo-gloss alignment and ID glossing--and through incremental baselines demonstrate consistent gains from each stage over fixed-pipeline approaches.(iii) We make the resulting curated data publicly available to support linguistically grounded SL research.

\section{Related Works}
\label{sec:related_works}
\subsection{Sign Language Understanding}
Computational research on SL understanding spans recognition \cite{li2020word, joze2018ms, sincan2020autsl,ozdemir2020bosphorussign22k}, translation \cite{zhou2023gloss, jiao2024visual, sincan2025gloss, kim2025leveraging, fish2025geosign}, production \cite{saunders2020progressive, walsh2024data, tang2025gloss, zuo2025signs} and assessment \cite{cory2024modelling, tarigopula2025posterior}. Early work focused on isolated sign recognition using handcrafted features, later replaced by deep learning approaches leveraging pose \cite{laines2023isolated, tunga2021pose}, video \cite{sincan2020autsl, zuo2023natural}, and multimodal embeddings \cite{hu2021signbert, zhao2023best, jiang2021skeleton}. Early studies on SL translation were largely gloss-based \cite{camgoz2020sign}. However, annotation is labor-intensive and prone to inconsistencies. To mitigate this, recent studies have proposed generating \textit{pseudo-glosses} from the spoken language sentences, which can improve translation performance while reducing reliance on manual annotation \cite{wong2024sign2gpt, asasi2025hierarchical, guo2025bridging}. Gong~\etal~\cite{gong2024llms} showed that LLMs can serve as effective sign language translators by leveraging in-context learning, while Inan~\etal~\cite{inan2025signalignlm} integrated multimodal sign language processing into LLMs through alignment objectives. Recent approaches target continuous sign language translation and production using large-scale sign language datasets \cite{li2025uni, uthus2023youtube}, by demonstrating strong gains from leveraging large data.

Linguistic studies provide essential insights into phonological structure \cite{jordan2015building, caselli2017asl, sehyr2021asl, ortega2025lexical}, such as handshape, movement, and location parameters, forming the base units of sign known as morphemes. While datasets with detailed linguistic annotation exist, they are small in scale, limiting usability in training deep models. To bridge this gap, a large-scale American Sign Language (ASL) dataset, ASL Citizen \cite{desai2023asl}, was collected by 52 signers taking its vocabulary from ASL-Lex \cite{caselli2017asl}, though substantial intra-class variability and label noise limit its suitability for sub-lexical modelling. Our approach directly targets this annotation bottleneck by introducing \textit{SignAgent}, an agentic framework specifically designed for sign-linguistic reasoning.

\subsection{Large Language Models}
Large Language Models (LLMs) \cite{brown2020language, achiam2023gpt, touvron2023llama, comanici2025gemini} have been extended beyond language understanding through tool use and Retrieval-Augmented Generation (RAG) \cite{lewis2020retrieval}. Structured retrieval approaches, like GraphRAG \cite{han2025retrieval}, further enhance reasoning by enabling models to operate over graph-based representations. Meanwhile, LLM-based agents \cite{xi2025rise,yao2022react} frame the model as a persistent decision-making entity that decomposes goals, calls tools, and iteratively refines plans based on environmental feedback, with dedicated reasoning mechanisms improving multi-step chain-of-thought performance \cite{yao2023tree, wang2024soft}.

Concurrently, multimodal LLMs extend their capabilities to other modalities, particularly vision \cite{li2024llava, achiam2023gpt}. Although promising, recent work has shown that they depend on weak visual encoders, are highly sensitive to prompt formulation, and lack the capacity to reliably capture fine-grained visual information.
Applications of LLMs to SL remain limited, with most work focusing on aligning text and visual features for translation \cite{zhou2023gloss, chen2024factorized, wong2024sign2gpt, sincan2025gloss}. A recent work \cite{kim2025leveraging} attempts to generate textual descriptions of SL videos, but LLMs often produce general or high-level textual descriptions rather than focusing on linguistic components. Our work builds on these directions by integrating multimodal grounding and reasoning within a unified agentic framework.

\begin{figure*}[t]
    \centering
    \includegraphics[width=1\linewidth]{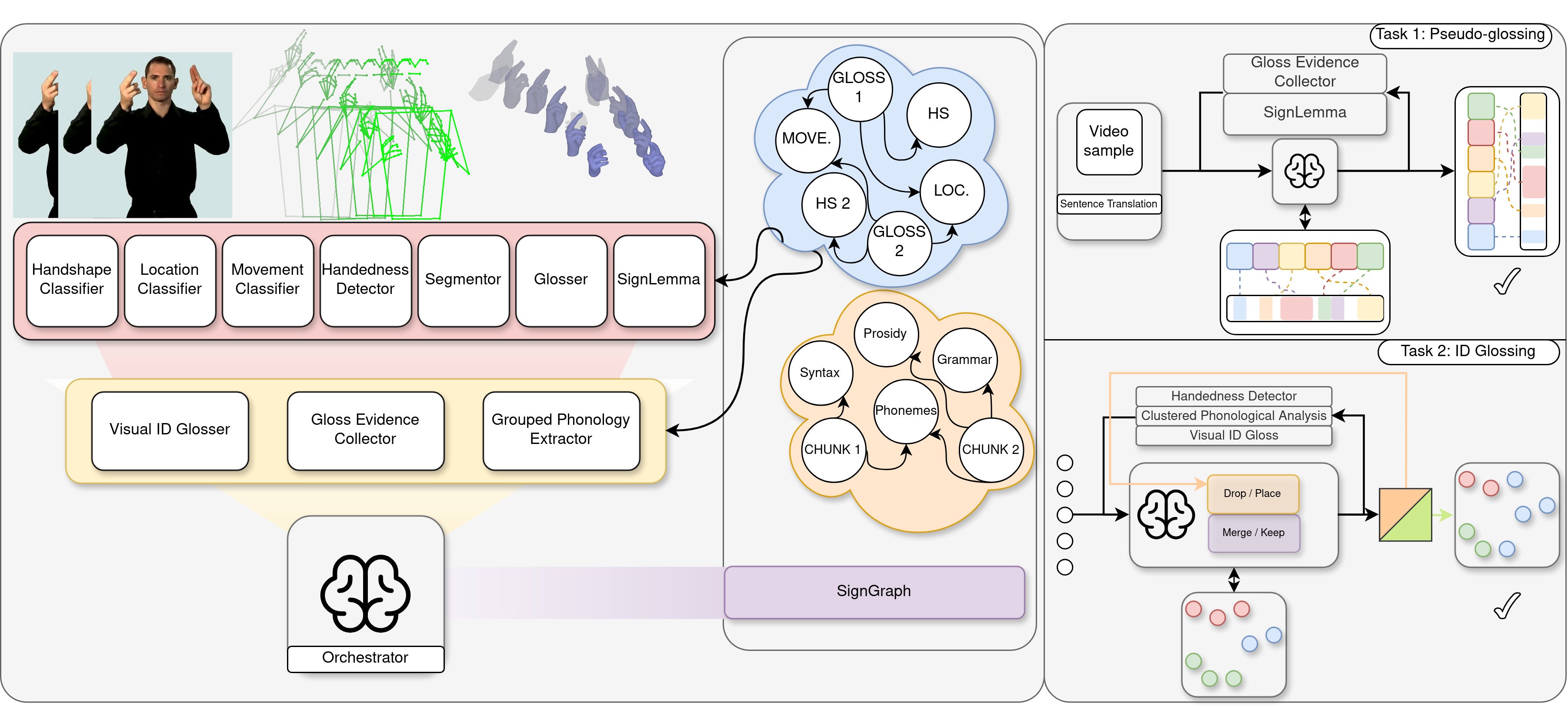}
    \caption{\emph{SignAgent} architecture (left) and pipeline for downstream tasks: pseudo-glossing and ID-glossing (right). The architecture consists of three core components. The \emph{SignAgent Orchestrator} is a reasoning Large Language model that acts as a linguistic expert, with access to a set of linguistic tools, and a knowledge retrieval agent, \emph{SignGraph} which retrieves \emph{lexical} and \emph{linguistic} data from \emph{knowledge graphs}, shown in blue and orange, respectively. The tools are assembled hierarchically, where \emph{Base Tools} (red), act to extract foundational SL information, across a range of modalities. The \emph{Enhanced Tools} (yellow) build upon the base tools to produce structured, task-ready evidence that the \emph{Orchestrator} can use in downstream tasks. }
    \label{fig:main}
\end{figure*}

\section{SignAgent}

\label{sec:method}
We present a novel agentic framework for SL video annotation and data curation (Fig. \ref{fig:main}). At its core is the \emph{SignAgent Orchestrator}, a reasoning LLM with tool-calling capabilities that can operate autonomously given input data and an initial prompt. The \emph{SignGraph} module is a hybrid language model tuned for retrieval-augmented generation, grounding the orchestrator through access to large-scale lexical and linguistic knowledge bases. Our framework comprises two hierarchically organised toolsets: \emph{Base Tools}, a collection of semantic, syntactic, and phonological modules that perform generalised analysis of SL; and \emph{Enhanced Tools}, which build upon and invoke the \emph{Base Tools} by integrating their outputs with subdomain-specific knowledge to enable enhanced and linguistically grounded analysis for downstream tasks.

\subsection{Agents}
Our method comprises two primary agents. The \emph{SignAgent Orchestrator} is a decoder-only large language model that acts as the central controller, managing reasoning, tool use, and information flow. \emph{SignGraph} operates alongside the \emph{Enhanced Tools} to provide linguistic grounding through retrieval and structured analysis of SL knowledge bases.

Given an input prompt \(x_0\), the orchestrator produces a sequence of intermediate reasoning states \(\{x_t, y_t, i_t\}_{t=0}^{n}\), where \(y_t\) denotes the model’s textual reasoning output at step \(t\), and \(i_t \in \{0, \dots, N\}\) represents either a tool invocation or a query to the knowledge graph. At each iteration \(t\), the orchestrator performs a ReAct-style \cite{yao2022react} reasoning loop:
\begin{equation}
x_t \;\rightarrow\; y_t \;\rightarrow\; i_t \;\rightarrow\; x_{t+1},
\end{equation}
where the model generates a reasoning trace \(y_t\), determines whether to execute a tool call or perform a graph-based retrieval via \emph{SignGraph}, and refines its internal state \(x_{t+1}\) from the result. Reasoning traces are structured through a JSON-based schema for consistent parsing. The process continues for \(n\) reasoning steps, constrained by a maximum of \(N\) invocations per run.

The combined \emph{SignGraph} and \emph{Enhanced Tools} modules provide linguistic grounding to the orchestrator by retrieving and synthesising structured lexical, phonological, and syntactic knowledge. Each retrieval or analysis operation expands the orchestrator’s context with linguistically informed representations, enhancing its ability to reason about SLs at multiple levels of abstraction. This design enables the system to integrate symbolic, statistical, and learned representations, supporting interpretable and informed decision-making across downstream SLs understanding tasks.

To provide linguistic and lexical grounding, we implement the \emph{LexicalKnowledgeGraph} and \emph{LinguisticKnowledgeGraph} as directed knowledge graphs. The \emph{LexicalKnowledgeGraph} contains nodes for lexical items (dictionary entries) and phonological components (handshape, movement, location), with edges encoding the relations between them. The \emph{LinguisticKnowledgeGraph} instead represents linguistic concepts extracted from books and other reference materials, with nodes such as core concepts and features, and edges (e.g., related-to) capturing the relationships between them. The \emph{SignGraph} retrieval agent reasons over both graphs, while the Enhanced Tools module directly consumes the lexical graph, as a dictionary, for downstream tasks such as phonology lookup and feature-based filtering.

\subsection{Base Tools}
Our framework implements a suite of \emph{Base Tools} (Fig.~\ref{fig:main}, red) for foundational, low-level linguistic analysis of SL videos, processing pose and video data to extract and classify key phonological components.
\begin{itemize}
    \item \textbf{Handshape Classifier:} Classifies handshape from regressed 3D hand keypoints  \cite{potamias2025wilor} via a k-nearest neighbour (k-NN) voting process on the resulting features against prototypes for X base and minor handshape.
    \item \textbf{Movement Classifier:} Extracts bilateral wrist trajectories from 3D pose data, constructs hybrid spectro-temporal features based on a Discrete Fourier Transform \cite{wang1984fast} and velocity statistics to capture dynamic properties of movement. Classification is carried out via k-NN with cosine distance.
    \item \textbf{Location Classifier:} Identifies signing location via body-relative keypoints for major locations, and prototypes matching for minor locations via k-NN, both with frame-level consensus voting.
    \item \textbf{Sign Segmentor:} Localises temporal gloss boundaries using a learned segmentation model \cite{11099255}.
    \item \textbf{Glosser:} Employs a state-of-the-art SL visual backbone trained on large-scale data \cite{wong2025signrep} to the task of dictionary retrieval via embedding distance nearest neighbour matching.
    \item \textbf{SignLemma:} Uses a text lemmatization technique \cite{wong2024sign2gpt} to convert spoken language text into plausible pseudo-gloss candidates.
    \item \textbf{Handedness Detector:} Infers per-sample handedness from segment-level hand detections, returning coarse labels and counts.
\end{itemize}
Technical specifications, mathematical formulations, and hyperparameter settings for these tools are provided in the Supplementary Material.

\section{Enhanced Tools}
\label{sec:enhanced-tools}
The \emph{Base Tools} (Sec.~3.2) provide fundamental, low-level outputs---visual embeddings, temporal segments, gloss candidates, and phonological predictions for handshape, movement, and location. The Enhanced Tools (Fig.~\ref{fig:main}, yellow) build on these signals to produce \emph{structured, task-ready evidence} that the \emph{SignAgent Orchestrator} can reason over (Sec.~3.1), closing the gap between per-component predictions and decision-ready hypotheses for downstream workflows such as pseudo-gloss reordering and ID glossing. Concretely, the \emph{Enhanced Tools} (i) fuse multi-modal cues (visual and phonological), (ii) summarise uncertainty via ranking and clustering, and (iii) expose compact, interpretable statistics (e.g., distances and overlap scores) that the agent can reason over to make linguistically grounded decisions. 

\noindent\textbf{Design principles.}
Each tool follows three principles:
\emph{coverage} (hybrid candidate generation to avoid single-source failure),
\emph{calibration} (scores that are comparable across segments/glosses),
and \emph{interpretability} (small sets of human-inspectable features and decisions).

\subsection{Gloss Evidence Collector}
\label{subsec:collector}
Given an input video, the Collector returns, for each segment, a ranked list of candidate glosses with calibrated relevance scores. It does not commit to a final label; instead, it assembles \emph{evidence} that the Orchestrator can weigh with contextual and linguistic constraints. We fuse visual retrieval with phonological matching. Visual candidates are obtained by calling the \emph{Sign Segmentor} to find gloss boundaries, followed by calling the \emph{Glosser}. %
Phonological candidates are added when component-level predictions from \emph{Handshape, Movement, and Location} classifiers agree with canonical lexical metadata.\footnote{Canonical labels come from the dictionary graph; details of label canonicalisation and handling of minor/base handshape variants are in the supplementary material.}

\paragraph{Hybrid candidate generation.}
For every gloss candidate $g$, extracted via the \emph{Glosser}, we query the dictionary graph to obtain its canonical phonology $g_c$, where c is the set of phonological components (e.g. handshape, location). The phonological classifiers provide top‑$k$ predictions $(\ell^c_r,p^c_r)$ for each component $c$, where $l$, $p$, and $^r$ are the label, confidence score, and rank respectively. We score visual-phonological agreement via:
\begin{equation}
s_{\text{phono}}(g)=\sum_{c} w_c \sum_{r=1}^{k} \frac{\mathbf{1}[\ell^c_r = g_c]}{r}\,p^c_r
\label{eq:s_phono}
\end{equation}
The nested sum rewards higher-confidence, better-ranked matches, and the factor $w_c$ averages over the number of canonical components the gloss has, such that $s_{\text{phono}}(g)$ ends up being the mean of those discounted matches. We then re-rank scores by combining the visual score with the phonological score, resulting in $s_{phono}$. Following this, we stack $s_{phono}$ alongside other visual features; forming a feature vector we use to train a lightweight Gradient Boosted Decision Tree (GBDT).

\paragraph{Output.}Per segment, we return ${(g_j,\hat y_j)}_{j=1}^M$, the top-$M$ candidates sorted by the learned relevance score $\hat y_j$ from the GBDT, alongside compact diagnostics (top margins, cross-modal agreement indicators). These scores are consumed by the Orchestrator in pseudo-gloss assignment and temporal reordering.

\subsection{Visual ID Glossing}
\label{subsec:visual-id}
Within a single semantic gloss category, we want to identify \emph{lexical variants} (ID glosses) by clustering video-level embeddings (\emph{SignRep} \cite{wong2025signrep}) derived from frame features. The result is a small number of visually coherent clusters that serve as a baseline partition prior to linguistic refinement.

\paragraph{Video representation.}
Given frame features, we compute a temporally averaged representation that is robust to variation in duration and speed. We adopt cosine distance and a thresholded, greedy assignment scheme: a sample joins the nearest cluster if the distance to its centroid is below $\tau$, otherwise it seeds a new cluster (as detailed in Supp C.4). This simple rule yields interpretable, reproducible clusters and a full pairwise distance matrix for analysis.

\paragraph{Output.}
For each gloss, the tool returns: cluster assignments, centroids, the distance matrix, nearest intra-/inter-cluster distances, and a flagged set of ``lone variants''. These structured artefacts feed the Orchestrator when weighing merges or deferring to linguistic evidence. 

\subsection{Clustered Phonological Analysis}
\label{subsec:phonology-id}
To complement visual clusters, we aggregate per-sample phonological predictions over clusters and quantify \emph{inter-cluster linguistic agreement}. This supports principled merging of visually separated but linguistically similar variants. 

\paragraph{Cluster-level overlap.}
For feature type $\phi$ (e.g. MANO hands, wrist features), let $P_\phi(C)$ be the set of labels appearing in the top-$k$ predictions (extracted via the \emph{Handshape, Movement and Location Classifiers}) of samples within cluster $C$. We compute Jaccard overlap
\begin{equation}
J_\phi(C_i,C_j) \;=\; \frac{\left| P_\phi(C_i) \cap P_\phi(C_j) \right|}{\left| P_\phi(C_i) \cup P_\phi(C_j) \right|},
\label{eq:jaccard}
\end{equation}
and define a merge recommendation when at least three feature types exceed a modest threshold, e.g., $\sum_{\phi}\mathbf{1}[J_\phi(C_i,C_j)\ge \tau_{\text{overlap}}] \ge 3$. Exact thresholds and ablations are provided in the Supplementary. When dictionary-level canonical labels are available, we additionally record cluster-level agreement with canonical phonology and expose this as an extra cue for merge decisions.

\paragraph{Output.}
For each cluster, we return a ranked list of candidate merge targets with (i) number of agreeing feature types, (ii) mean overlap score, and (iii) strongest agreeing properties (e.g., ``handshape-base, movement''). This targeted view reduces search space while retaining transparency.

\section{Experimental Setup}
\label{sec:experiments}
To assess our agent we evaluate \emph{SignAgent} on controlled annotation tasks that test whether the agent can integrate multi–modal linguistic evidence and produce structured, auditable outputs. Across all experiments, the agent interacts with a set of callable tools (Sec.~\ref{sec:enhanced-tools}) and reasons over their structured responses. 

We evaluate our framework on both British Sign Language (BSL) and American Sign Language (ASL) datasets. For pseudo-gloss annotation (\cref{sec:pseudogloss}) we choose BSLCorpus \cite{schembri2013building}; a continuous SL dataset with ground truth manually annotated gloss labels. For testing, we derive a hard set from the test partition, where traditional Pseudo-gloss annotations struggle. 
For our BSL lexical dictionary, we choose BSLSignbank \cite{jordan2014bsl}, a smaller but phonologically annotated dictionary resource containing 3451 individual IDs. We use the video data for retrieval and the corresponding phonological metadata to build our knowledge graph.
For ID-glossing (\cref{sec:idgloss}), we use ASLCitizen \cite{desai2023asl}; a large-scale in-the-wild dataset containing aligned but phonologically distinct variants of glosses. Specifically, we choose samples from the test set that have corresponding codes in ASL-Lex \cite{caselli2017asl}. As such we choose ASL-Lex2.0 \cite{sehyr2021asl} as our ASL Dictionary, and use 1144 samples from this.

\section{Task 1: Pseudo-gloss Annotation}
\label{sec:pseudogloss}
Given a translated text sentence and its corresponding signed-video segment, the agent must infer the appropriate gloss labels and place them in a well-ordered sequence aligned to the video. The pipeline for this pseudo-gloss annotation is illustrated in Fig.\ref{fig:main} (top-right). The agent first queries \emph{SignLemma} over the text to obtain an initial set of gloss candidates. After this point, it does not invent new glosses; instead, it reorders and selects among these candidates by weighing multimodal evidence gathered through tool calls, including sign segmentation, visual retrieval, phonological similarity, and activity-based temporal cues.

\subsection{Workflow and Tool Calls}
\label{sec:pseudogloss:workflow}
\paragraph{Controller.}
The agent alternates between (i) LLM reasoning steps and (ii) executable tool calls that return a structured JSON. It maintains a compact state containing the input sample text sentence, the token set $T$, and per-segment evidence. A hard cap on the number of tool invocations for reproducibility. Low-level implementation details are in the supplementary material.

\paragraph{Tool-calling protocol.}
We restrict the agent to a small, auditable API.

\begin{enumerate}
  \item \emph{SignLemma}: returns the normalized token set $T=\{t_1,\dots,t_M\}$ that must be assigned exactly once. This call defines the assignment constraints for the episode.
  \item \emph{GlossEvidenceCollector}:
  returns, for each segment $s\in S$, (a) temporal bounds, (b) visual activity diagnostics (e.g., hand detections, wrist keypoint dynamics), and (c) a top-$k$ list of candidate dictionary glosses with visual embedding similarities, per-component phonological predictions (handshape, movement, location) and learned relevence score. The agent may vary $k$ across at most two calls to probe candidate stability.
\end{enumerate}

\paragraph{Reasoning cues and assignment criteria.}
The assignment problem is defined over the segment set \(S\) returned by \emph{Gloss Evidence Collector} (which obtains temporal sign boundaries from the segmentor) and the multiset of pseudo-gloss tokens \(T = \{t_1,\dots,t_M\}\) normalised by \emph{SignLemma}. For each \(t_i \in T\), the agent scores candidate pairs \((s,c)\) with segments \(s \in S\) and dictionary candidates \(c \in C(s)\), and selects
\[
(s^\star, c^\star) = \arg\max_{s \in S,\, c \in C(s)} \mathrm{score}(t_i, s, c;\text{context}),
\]
where the score aggregates five evidence types:
\begin{itemize}
  \item \emph{Visual similarity}: cosine similarity between the segment embeddings and each dictionary candidate's embeddings (from \emph{Gloss Evidence Collector})
  \item \emph{Phonological overlap}: agreement between per-segment predictions (handshape, movement, location) and a candidate's canonical properties via inverse dictionary lookup on $t_i$ (from \emph{SignLemma}). We use the discounted, confidence-weighted overlap described in Sec.~\ref{sec:enhanced-tools}.
  \item \emph{Hand activity}: aggregate hand detections and wrist activity metrics (from \emph{Gloss Evidence Collector}) used to downweight candidates in weakly active signing segments.
  \item \emph{Temporal coherence}: segment duration and position in the signing sequence, favoring plausible placements (e.g., short function-like tokens appearing in brief, isolated segments).
  \item \emph{Semantic context}: alignment between the reference sentence and candidate gloss semantics exposed by \emph{SignLemma}.
\end{itemize}
High-confidence visual anchors seed the mapping; ambiguous cases are resolved using phonological and temporal cues. The final pseudo-gloss sequence is obtained by reading off tokens in the temporal order of their assigned segments, subject to token-conservation and no-external-token constraints.

\paragraph{Validation.}
We verify the conservation and no-hallucination constraints by set equality over input/output tokens:
\begin{equation}
\texttt{valid} \;\Leftrightarrow\; (T_{\text{output}} = T_{\text{input}})\;\wedge\; |T_{\text{output}}|=|T_{\text{input}}|.
\label{eq:validation}
\end{equation}
Invalid episodes are rejected rather than corrected, ensuring sequence metrics reflect \emph{reordering} quality rather than insertion/deletion.

\paragraph{Output.}
For each input sentence, the agent returns a structured pseudo-gloss annotation rather than free-form text (Supp.~Fig.~S.1). The output consists of the final ordered sequence of gloss tokens aligned to the video, together with a record of which token is assigned to which segment and which dictionary candidate it realises, along with a short justification based on the available evidence. Because the controller enforces token conservation and forbids the introduction of new tokens, this sequence is a faithful reordering of the initial pseudo-gloss hypotheses. The accompanying alignment record allows us to compute per-token and per-segment metrics (e.g.\ token accuracy, segment-level precision/recall, and sequence edit distance) while keeping every assignment auditable.

\subsection{Results}
\label{sec:pseudo_gloss_results}
We evaluate the predicted pseudo-gloss sequences against the ground-truth gloss set in BSLCorpus, mapping both reference and hypothesis glosses to normalised word forms (lowercased English lemmas).
We report two complementary sequence metrics: the longest common subsequence (LCS) as a percentage of the reference sequence length,  measuring in-order token alignment; 
and Kendall’s $\tau$ rank correlation over the positions of matched gloss types, which captures how well the relative ordering of tokens is preserved ($\tau=1$ is perfect agreement, $\tau=0$ is uncorrelated, and $\tau<0$ indicates systematic inversions).

Table \ref{tab:pseudo_gloss_alignment} is structured as an incremental ablation through the baseline progression:
\emph{Sign2GPT Lemmatization $\rightarrow$ GBDT+fuzzy $\rightarrow$ SignAgent}.
Each step introduces a specific capability while holding others fixed. In particular, \emph{GBDT+fuzzy} corresponds to \emph{SignAgent without the LLM-based orchestrator reasoning}, whilst retaining the same multimodal evidence sources (SignRep visual retrieval and phonological classifiers) and token-constrained assignment formulation. The consistent improvements, therefore, isolate the contribution of (i) multimodal evidence aggregation and learned ranking, and (ii) agentic LLM reasoning for resolving conflicting evidence.

As summarised in Table~1, the \emph{Sign2GPT Lemmatization} baseline performs reasonably on Fair (57.26\% LCS, $\tau=0.232$) but poorly on Poor (34.52\% LCS, $\tau=-0.333$). Adding a GBDT scorer with fuzzy semantic matching gives modest, consistent gains (58.00\% / 0.297 on Fair, 38.69\% / $-0.083$ on Poor, 55.97\% / 0.257 combined). Our full \emph{SignAgent} achieves the best results (60.85\% / 0.374 on Fair, 47.02\% / 0.083 on Poor, 59.40\% / 0.343 combined), yielding absolute improvements of 4.53 LCS points and 0.17 Kendall-$\tau$ over lemmatisation (3.43 / 0.086 over GBDT) and removing the negative rank correlation on the hardest sentences, indicating that agentic, tool-augmented reasoning over multi-modal evidence better resolves difficult reordering decisions than fixed feature-based baselines.

\begin{table}[t]
\centering
\setlength{\tabcolsep}{3.5pt}
\renewcommand{\arraystretch}{0.95}
\begin{tabular}{@{}l cc cc cc@{}}
\toprule
& \multicolumn{2}{c}{\textbf{Fair}}
& \multicolumn{2}{c}{\textbf{Poor}}
& \multicolumn{2}{c}{\textbf{Combined}} \\
\cmidrule(lr){2-3} \cmidrule(lr){4-5} \cmidrule(lr){6-7}
\textbf{Method}
  & LCS\% $\uparrow$ & $\tau$ $\uparrow$
  & LCS\% $\uparrow$ & $\tau$ $\uparrow$
  & LCS\% $\uparrow$ & $\tau$ $\uparrow$ \\
\midrule
Guo~\etal Lemma.$^\dagger$~\cite{guo2025bridging}
   & 23.98 & 0.280 & 14.78 & $-$0.333 & 23.02 & 0.216 \\
\rowcolor{black!5}
Sign2GPT Lemma.~\cite{wong2024sign2gpt}
  & 57.26 & 0.232 & 34.52 & $-$0.333 & 54.87 & 0.173 \\
GBDT + fuzzy
  & 58.00 & 0.297 & 38.69 & $-$0.083 & 55.97 & 0.257 \\
\rowcolor{black!5}
\textbf{SignAgent} {\small(Ours)}
  & \textbf{60.85} & \textbf{0.374}
  & \textbf{47.02} & \textbf{0.083}
  & \textbf{59.40} & \textbf{0.343} \\
\bottomrule
\end{tabular}
\caption{Pseudo-gloss alignment results. Metrics are averaged per subset (Fair, Poor, Combined). Combined is averaged across both Fair and Poor. The GBDT baseline uses the same multimodal features as SignAgent but replaces the LLM orchestrator with a learned ranker. $^\dagger$Reproduced using published prompts with GPT-OSS:120B \cite{achiam2023gpt}}
\label{tab:pseudo_gloss_alignment}
\end{table}

\section{Task 2: ID Glossing}
\label{sec:idgloss}
We evaluate whether the agent can \emph{refine within-gloss clusters into stable ID glosses} by reasoning over structured, multi–modal evidence returned by tools (Fig. \ref{fig:main} bottom-right). Starting from a visual baseline partition, the agent proposes \textsc{MERGE}/\textsc{KEEP} operations that are justified by (i) inter/intra–cluster visual distances, (ii) cluster–level phonological agreement, and (iii) handedness compatibility. The entire procedure is auditable: every operation cites the exact fields returned by the tools. Please see supplementary material for details.

\subsection{Workflow and Tool Calls}
\label{sec:idgloss:workflow}
\paragraph{Controller.}
As in Section~\ref{sec:pseudogloss:workflow}, the agent runs a \emph{reason–act} loop that alternates LLM reasoning with executable tool calls that return JSON. The persistent state holds the baseline visual clusters for a given gloss, tool outputs (distance matrices, overlap tables, handedness labels), and a log of proposed operations.

\paragraph{Tool-calling protocol.}
\begin{enumerate} 
  \item \emph{Visual ID Glossing}: returns baseline clusters, centroids, full pairwise distance matrix $D$, nearest intra/inter–cluster distances, and flagged singletons/lone variants. These constitute the initial partition and quantitative visual evidence.
  \item \emph{Clustered Phonological Analysis}: aggregates per–sample phonological predictions over clusters and provides ranked merge recommendations with feature–type Jaccard overlaps $J_\phi(C_i,C_j)$ and counts of agreeing feature types. Canonical properties of dictionary entry for given gloss is returned for calibrated agreement.
  \item \emph{Handedness Detector}: returns per-sample left/right/both/mixed labels and one-/two-handed flags, alongside aggregate counts and left/right ratios.

\end{enumerate}
Tools returns (a) concise recommendations to focus the search and (b) detailed numerical context (distances, overlaps, feature–level tables) to support verification. The controller first narrows candidates using recommendations; then validates them against the detailed fields before committing any operation. 

\paragraph{Reasoning Cues.} For candidate clusters $(C_i,C_j)$ the agent computes a composite decision score from:
\begin{itemize}
  \item \emph{Visual proximity}: inter–cluster distance statistics from \emph{Visual ID Gloss}, using $D$ and nearest inter/intra distances to gauge separability.
  \item \emph{Phonological overlap}: number of feature types $\phi$ whose Jaccard $J_\phi(C_i,C_j)$ exceeds $\tau_{\text{overlap}}$, plus mean overlap; canonical agreement counters, from \emph{Clustered Phonological Analysis}
  \item \emph{Handedness compatibility}: uses \emph{Handedness Detector} outputs to rule out implausible merges and to treat mirror variants as mergeable; cluster-level handedness proportions help disambiguate boundary merges.
\end{itemize}

Thresholds shared in prompts are treated as \emph{guidelines}, not hard rules; the agent reasons holistically over the returned numbers and justifies any deviation in the operation log.

\paragraph{Cluster refinement operations.}
Given an initial tool call to \emph{Visual ID Glossing}, the agent applies two auditable operations to the visual baseline clusters:
\begin{description}
  \item[\textsc{MERGE}.] Reassign samples by (i) moving singletons to multi–member clusters, or (ii) merging multiple clusters. Heuristics encourage merges when inter–cluster distance is small relative to the baseline threshold and when $\geq\!3/5$ (singletons) or $\geq\!4/5$ (multi–cluster) phonological feature types agree above a modest overlap, provided handedness is compatible. All decisions cite $D$, the per–feature $J_\phi$ table, and handedness fields.
  \item[\textsc{KEEP}.] Preserve separation when evidence is insufficient or conflicting (e.g., large $D$ gap, weak phonology agreement, incompatible handedness), explicitly logging the blocking statistic(s). 
\end{description}

\paragraph{Validation and correction stage.}
After the initial refinement, the system validates that every sample is assigned \emph{exactly once}. Let $S$ be the set of baseline sample keys and $A$ the multiset of assigned samples in the agent’s output. We require:
\begin{equation}
\texttt{valid}\;\Leftrightarrow\; (A=S)\;\wedge\;\big(\forall s\in A:\,\mathrm{count}(s)=1\big).
\label{eq:id-valid}
\end{equation}

If validation fails, the agent receives a single constrained correction pass that may reassign duplicates or place missing samples into compatible
clusters, but may not alter the global cluster topology. Uncorrectable
outputs are flagged for post-hoc analysis, preserving the full reasoning
trace.

\paragraph{Output.}
For each gloss, the agent returns a structured
ID-glossing annotation as a schema-validated JSON record (Supp.~Fig.~S.2). The output contains:
(i)~the final refined cluster partition, where each cluster includes assigned samples and a justification citing the visual distance matrix ${D}$, per-feature Jaccard overlaps $J_\phi$, and handedness statistics;
(ii)~a singleton-review log recording whether each baseline singleton was kept or merged with supporting evidence;
(iii)~phonological adjustment records specifying operations and rationale; and (iv)~summary statistics, including a global confidence score $c \in [0,1]$ and a validation-status flag verifying Eq.~(\ref{eq:id-valid}). Every MERGE/KEEP decision remains auditable at the level of individual samples and clusters.

\subsection{Results}
\begin{figure}[h]
        \centering
        \includegraphics[width=.8\linewidth]{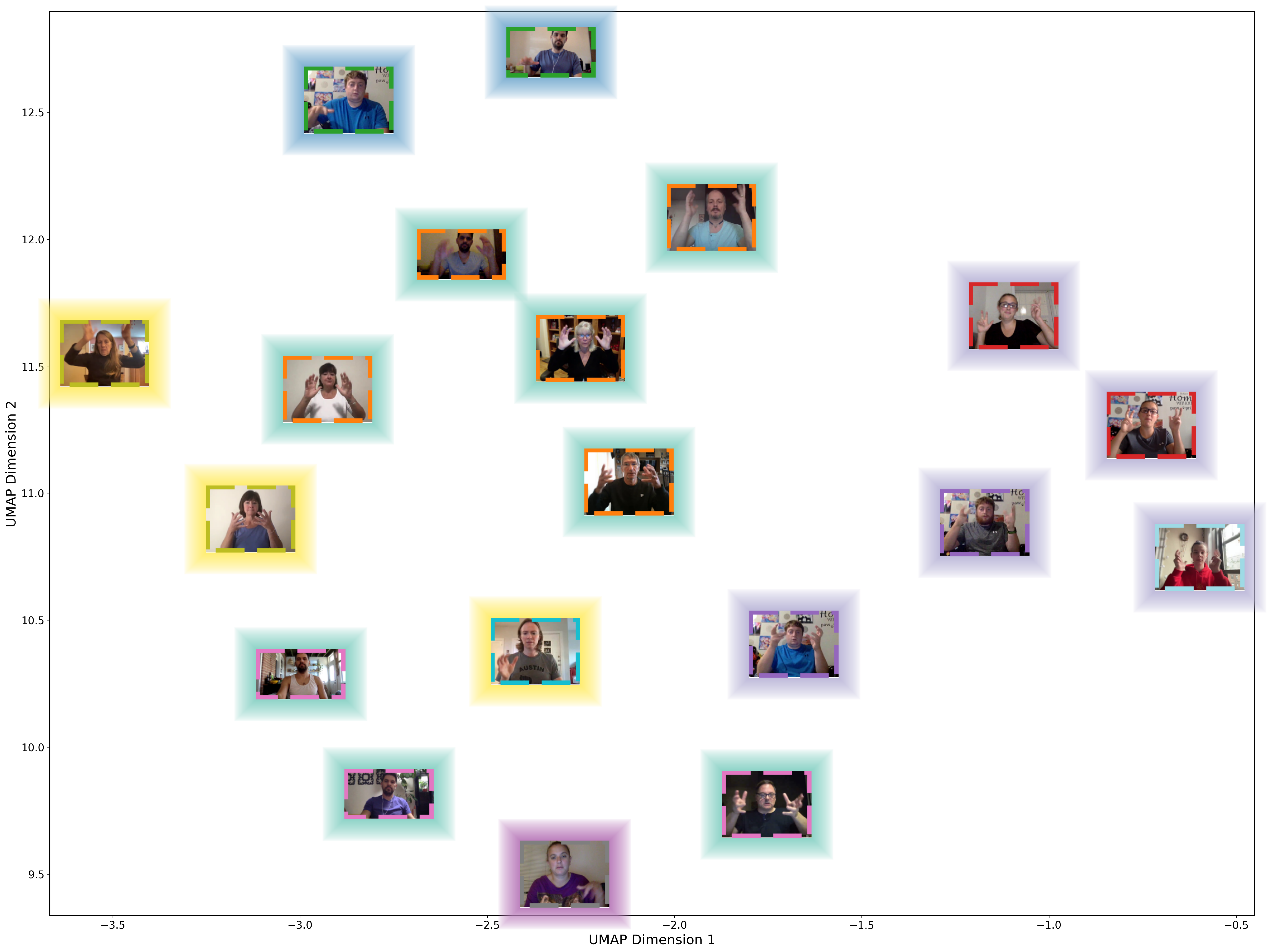}
        \caption{UMAP of SignRep sign embeddings for the gloss `Basketball'. Baseline SignRep clusters are shown by the dashed borders (one colour per cluster). The glowing borders denotes \emph{SignAgent} (Ours) assigned clusters.}
        \label{fig:figure2}
\end{figure}

\begin{table}[t]
    \centering
    \footnotesize
    \setlength{\tabcolsep}{2.2pt}
    \begin{tabular}{cccccc}
        \toprule
        \textbf{Method} & \textbf{IDs/gloss} & \textbf{Total IDs} & \textbf{$H$ (bits)$\downarrow$} & \textbf{Silhouette$\uparrow$} & \textbf{Calinski--H$\uparrow$} \\
        \midrule
        \makecell[c]{SignRep  \small{(Baseline)}} & 4.81 & 5431 & 1.910 & -0.0402 & 6.75 \\
        \rowcolor{black!5} \makecell[c]{\textbf{SignAgent} \small{(Ours)}} & 2.30 & 2602 & \textbf{0.764} & \textbf{0.0582} & \textbf{7.58} \\
        \bottomrule
    \end{tabular}
    \caption{Within-gloss clustering metrics comparing SignRep \cite{wong2025signrep} and \emph{SignAgent} cluster assignment calculated by embedding cluster samples using SkeletonVAE model \cite{cory2024modelling}, trained purely on pose data as to avoid bias towards RGB or baseline. $H$ (bits) is the entropy of the cluster-size distribution per gloss (lower indicates fewer, less fragmented clusters), Silhouette is the mean silhouette coefficient over all instances (higher indicates better separation and compactness), 
    and Calinski--H is the Calinski--Harabasz ratio of between- to within-cluster variance (higher indicates better-defined clusters).}
    \label{tab:within_gloss_clustering}
\end{table}
\paragraph{Quantitive Results: } We evaluate clustering performance for ID glossing against SignRep, a state-of-the-art vision encoder adapted for ID glossing across 1,130 glosses. As shown in Table~\ref{tab:within_gloss_clustering}, SignAgent yields substantially less fragmented within-gloss structure (2.3 vs.\ 4.8 IDs/gloss, lower cluster-size entropy) while also improving cluster quality (silhouette $-0.04 \rightarrow 0.06$, Calinski--Harabasz $6.75 \rightarrow 7.58$), indicating fewer, more coherent clusters than SignRep \cite{wong2025signrep}.
\paragraph{Qualitative Results:}
\cref{fig:figure2} demonstrates \emph{SignAgent}'s stronger clustering performance when compared to the \emph{SignRep} baseline. A notable result here is the right-hand purple cluster. Whilst SignRep failed to group the ground truth cluster due to visual variances in the embedding space, SignAgent successfully merged candidate clusters which share identical phonological features. 
\paragraph{Expert-Grounded Cluster Consistency:}
ID-gloss refinement is grounded in expert-curated lexical resources.
Cluster-level phonological agreement is evaluated against canonical
handshape, movement, and location properties from ASL-Lex2.0 \cite{sehyr2021asl} via SignGraph retrieval. Thus, merge decisions are constrained by established linguistic metadata rather than purely visual similarity. Reductions in cluster fragmentation and improvements in separation metrics are achieved under these expert-defined phonological constraints, providing structured alignment with established lexical distinctions.

\section{Conclusion}
\label{sec:disc_conc}
We framed sign language annotation as an agentic reasoning problem and showed that a structured LLM scaffold can materially improve both accuracy and interpretability. Across pseudo-gloss annotation and ID glossing, SignAgent—combining an LLM orchestrator with linguistic tools and SignGraphRAG—outperforms rule-based and learned baselines while keeping every decision grounded in explicit evidence. In pseudo-gloss annotation, the agent consistently improves ordering and sequence metrics, especially on challenging cases where conventional methods fail. For ID glossing, jointly
reasoning over visual distances, phonological overlap, and
handedness produces fewer, more coherent clusters than purely visual baselines.
Although SignAgent offers a scalable and linguistically grounded approach to sign language annotation, several limitations remain.
The framework still depends on existing lexical resources and tools, and it only partially captures non-manual and prosodic structure. Extending SignAgent to low-resource sign languages, enriching the toolset with non-manual and morpho-syntactic analyses, and exploring joint optimisation of tools and controller are important next steps. SignAgent is intended as an assistive tool for linguists and dataset curators, not as a replacement for expert linguistic judgement; our results suggest that agentic LLMs can serve as scalable, auditable collaborators for sign language annotation.

\section*{Acknowledgements}
This work was supported by the the Innosuisse IICT Flagship (PFFS-21-47), EPSRC grant APP24554 (SignGPT-EP/Z535370/1), and through funding from Google.org via the AI for Global Goals scheme. This work reflects only the author’s views and the funders are not responsible for any use that may be made of the information it contains.

\bibliographystyle{splncs04}
\bibliography{main}
\end{document}